\documentclass[11pt,a4paper]{article}
\usepackage[hyperref]{acl2018}
\usepackage{times}
\usepackage{latexsym}
\usepackage{amsfonts}
\usepackage{amsmath}
\usepackage{xcolor}

\usepackage{url}
\usepackage{graphicx}
\usepackage{tabularx}
\usepackage{enumitem}

\newcommand{\tabhline}[0]{\\[0.4ex]\hline\\[-2.5ex]}

\newcommand{\densesection}[1]{\vspace*{-0.4ex}\section{#1}\vspace*{-0.35ex}}
\newcommand{\densesubsection}[1]{\vspace*{-0.375ex}\subsection{#1}\vspace*{-0.2ex}}

\aclfinalcopy 

\title{Multi-task learning for Joint Language Understanding and Dialogue State Tracking}

\author{Abhinav Rastogi \\
  Google AI \\
  Mountain View \\
  {\tt abhirast@google.com} \\\And
  Raghav Gupta \\
  Google AI \\
  Mountain View \\
  {\tt \, \, raghavgupta@google.com} \\\And
  Dilek Hakkani-Tur \\
  Google AI \\
  Mountain View \\
  {\tt dilek@ieee.org} \\}

\date{}

\begin{document}
\maketitle

\begin{abstract}

This paper presents a novel approach for multi-task learning of language
understanding (LU) and dialogue state tracking (DST) in task-oriented dialogue
systems. Multi-task training enables the sharing of the neural network layers
responsible for encoding the user utterance for both LU and DST and improves
performance while reducing the number of network parameters.
In our proposed framework, DST operates on a set of
candidate values for each slot that has been mentioned so far. These candidate
sets are generated using LU slot annotations for the current user utterance,
dialogue acts corresponding to the preceding system utterance and the dialogue
state estimated for the previous turn, enabling DST to handle slots with a
large or unbounded set of possible values and deal with slot values not seen
during training. Furthermore, to bridge the gap between training and inference,
we investigate the use of scheduled sampling on LU output for the current user
utterance as well as the DST output for the preceding turn.
\end{abstract}

 \densesection{Introduction}
Task-oriented dialogue systems interact with users in natural language to
accomplish tasks they have in mind, by providing a natural language interface to
a backend (API, database or service). State of the art approaches 
to task-oriented dialogue systems typically consist of a language understanding
(LU) component, which estimates the semantic parse of each user utterance and
a dialogue state tracking (DST) or belief tracking component, which keeps track
of the conversation context and the dialogue state (DS). Typically, DST uses the semantic
parse generated by LU to update the DS at every dialogue turn. The DS
accumulates the preferences specified by the user over the dialogue and is used
to make requests to a backend. The results from the backend and the dialogue
state are then used by a dialogue policy module to generate the next system
response.


\begin{figure}[t]
\small
  \begin{tabular}{ l l}
\textbf{System:} & Hello! How can I help?\\
 Acts: & \textit{greeting}\\
\textbf{User:} & Hello, book me a table for two at Cascal.\\
 Intent: & RESERVE\_RESTAURANT\\
 Acts: & \textit{greeting},  \textit{inform}(\#people), \textit{inform}(restaurant)\\
 State: & \texttt{restaurant=Cascal,\#people=two}\\\\
\textbf{System:} & I found a table for two at Cascal at 6 pm.\\
  & Does that work?\\
 Acts: & \textit{offer}(time=6 pm)\\
\textbf{User:} & 6 pm isn't good for us. How about 7 pm?\\
Acts: & \textit{negate}(time), \textit{inform}(time)\\
State: &\texttt{restaurant=Cascal,\#people=two,}\\
&\texttt{time=7 pm}\\
  \end{tabular}
  \caption{A dialogue with user intent, user and system dialogue acts,
  and dialogue state.}
  \label{fig:dialogue-example}
\end{figure}

Pipelining dialogue system components often leads to error propagation, hence
joint modeling of these components has recently gained popularity
\cite{henderson2014, mrkvsic2017neural, BingLiu2017}, owing to computational
efficiency as well as the potential ability to recover from errors introduced
by LU. However, combining joint modeling with the ability to scale
to multiple domains and handle slots with a large set of possible values,
potentially containing entities not seen during training, are
active areas of research.


In this work, we propose a single, joint model for LU and DST trained
with multi-task learning. Similar to~\citealt{BingLiu2017}, our model employs a
hierarchical recurrent neural network to encode
the dialogue context. Intermediate feature representations from this network are
used for identifying the intent and dialogue acts, and tagging slots in the user
utterance. Slot values obtained using these slot tags (as shown in Figure
\ref{fig:clu-example}) are then used to update the set of candidate values
for each slot. Similar to ~\citealt{rastogi2017}, these candidate values are then
scored by a recurrent scoring network which is shared across all slots, thus
giving an efficient model for DST which can handle new entities that are not
present in the training set - i.e., out-of-vocabulary (OOV) slot values.

\begin{figure}[t]
\small
  \begin{tabular}{ l c c c c c c}
\textbf{Utterance:} & Table & for & two & at & Olive & Garden\\
& $\downarrow$ & $\downarrow$ & $\downarrow$ & $\downarrow$ & $\downarrow$ & $\downarrow$ \\
\textbf{Slot Tags:} & O & O & B-\# & O & B-rest & I-rest\\
\end{tabular}
\caption{IOB slot tags for a user utterance. Slot values
  \textit{\# = two} and \textit{rest = Olive Garden} are obtained from corresponding B and I tags.}
  \label{fig:clu-example}
\end{figure}

During inference, the model uses its own predicted slot tags and
previous turn dialogue state. However, ground truth slot tags and
dialogue state are used for training to ensure stability. Aiming to bridge
this gap between training and inference, we also propose a
novel scheduled sampling ~\cite{bengio2015scheduled} approach to joint language
understanding and dialogue state tracking.

The paper is organized as follows: Section \ref{sec:related-work} presents
related work, followed by Section \ref{sec:dialog-encoder}
describing the architecture of the dialogue encoder, which encodes the
dialogue turns to be used as features by different tasks
in our framework. The section also defines and outlines the implementation of
the LU and DST tasks. Section
\ref{sec:scheduled} describes our setup for scheduled sampling. We then
conclude with experiments and discussion of results.

 \densesection{Related Work}
\label{sec:related-work}
The initial motivation for dialogue state tracking came from the
uncertainty in speech recognition and other sources
\cite{williams2007partially}, as well as to provide a
comprehensive input to a downstream dialogue policy component deciding the next
system action. Proposed belief tracking models have
ranged from rule-based \cite{wang2013simple}, to
generative \cite{thomson2010bayesian}, discriminative \cite{henderson2014},
other maximum entropy models
\cite{williams2013multi} and web-style ranking \cite{williams2014web}.

Language understanding has commonly been modeled as a combination of intent and
dialogue act classification and slot tagging \cite{tur2011spoken}. Recently,
recurrent neural network (RNN) based approaches have shown good results for LU.
\citealt{hakkani2016multi} used a joint RNN for intents, acts and slots to achieve
better overall frame accuracy. In addition, models such as \citealt{chen2016end},
\citealt{bapna2017sequential} and \citealt{vivian2017} further improve LU results by
incorporating context from dialogue history.

\citealt{henderson2014} proposed a single joint model for
single-turn LU and multi-turn DST to improve belief tracking performance.
However, it relied on manually constructed semantic dictionaries to identify
alternative mentions of ontology items that vary lexically or morphologically.
Such an approach is not scalable to more complex domains ~\cite{mrkvsic2017neural}
as it is challenging to construct semantic dictionaries that can cover all
possible entity mentions that occur naturally in a variety of forms in natural
language. \citealt{mrkvsic2017neural} proposed the NBT model which
eliminates the LU step by directly operating on the user utterance. However, their
approach requires iterating through the set of all possible values for a slot,
which could be large or potentially unbounded (e.g. date, time,
usernames). \citealt{perez2017dialog} incorporated end-to-end memory
networks, as introduced in \citealt{sukhbaatar2015end}, into state tracking
and \citealt{BingLiu2017} proposed an end-to-end model for belief tracking.
However, these two approaches cannot accommodate OOV slot values
as they represent DS as a distribution over all
possible slot values seen in the training set.

To handle large value sets and OOV slot values, ~\citealt{rastogi2017} proposed an
approach, where a set of value
candidates is formed at each turn using dialogue context.
The DST then operates on this set of candidates. In this work, we
adopt a similar approach, but our focus is on joint modeling of LU and
DST, and sampling methods for training them jointly.



 \densesection{Model Architecture}
\label{sec:dialog-encoder}
\begin{figure*}[t]
  \centering
  \includegraphics[width=\linewidth]{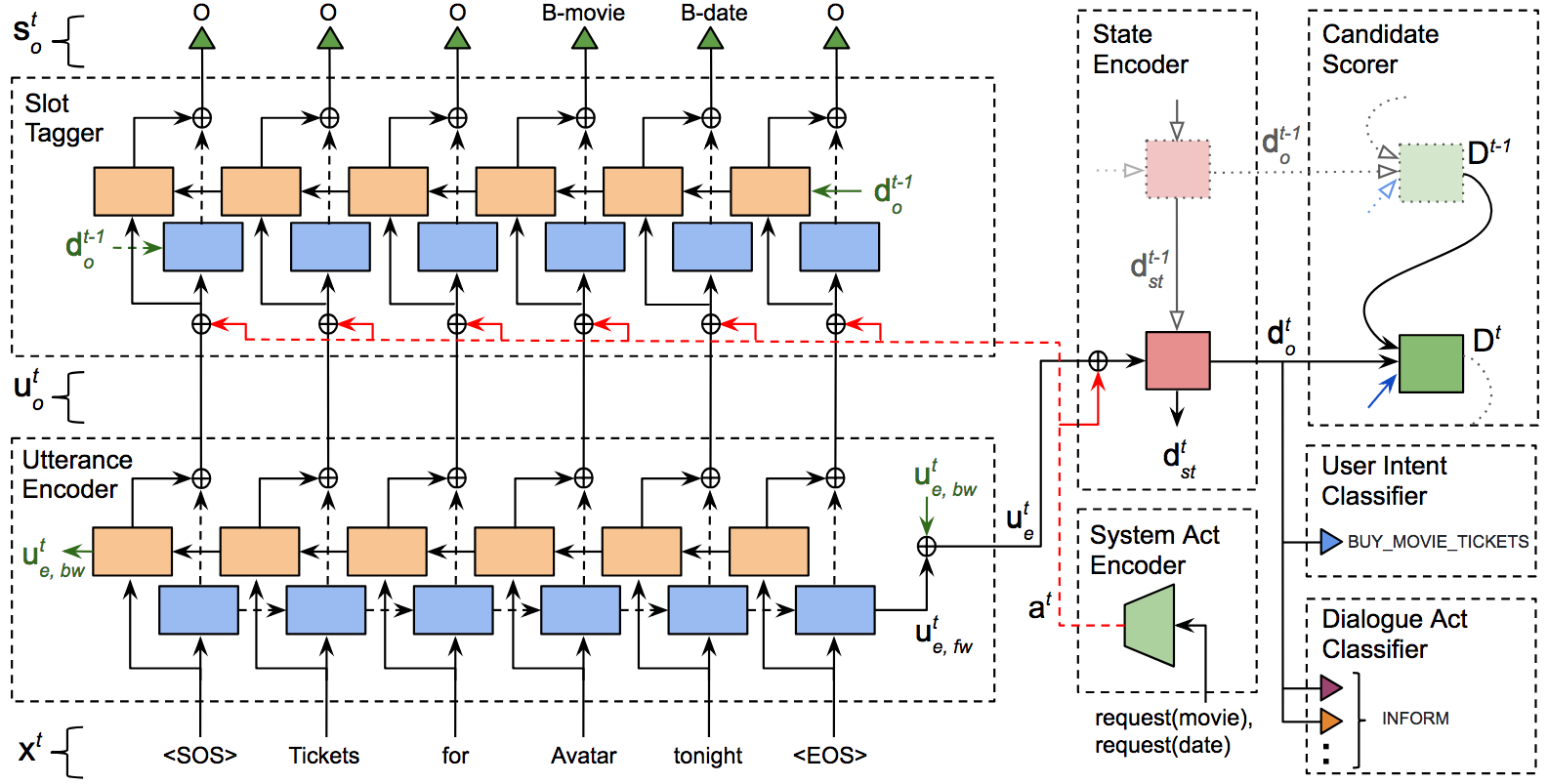}
  \caption{Architecture of our joint LU and DST model as described in Section
    \ref{sec:dialog-encoder}. $x^t$ is the sequence of user utterance token
    embeddings, $a^t$ is the system act encoding and blue arrows indicate
    additional features used by DST as detailed in Section
    \ref{sec:candidate-scorer}.}
  \label{fig:fig2}
\end{figure*}
Let a dialogue be a sequence of $T$ turns, each turn containing a user
utterance and the preceding system dialogue acts output by the dialogue manager.
Figure \ref{fig:fig2} gives an overview of our model architecture, which
includes a user utterance encoder, a system act encoder, a state encoder, a slot tagger
and a candidate scorer. At each
turn $t \in \{1,...,T\}$, the model takes a dialogue turn and the
previous dialogue state
$D^{t-1}$ as input and outputs the predicted user intent, user dialogue acts,
slot values in the user utterance and the updated dialogue state $D^t$.

As a new turn arrives, the system act encoder (Section \ref{sec:system-act})
encodes all system dialogue acts in the turn to generate the system dialogue act vector
$a^t$. Similarly, the utterance encoder (Section \ref{sec:utterance-encoder})
encodes the user utterance into a vector $u_e^t$, and also generates contextual
token embeddings $u_o^t$ for each utterance token. The state
encoder (Section \ref{sec:state-encoder})
then uses $a^t$, $u_e^t$ and its previous turn hidden state, $d_{st}^{t-1}$, to generate the
dialogue context vector $d_o^t$, which summarizes the entire
observed dialogue, and its updated hidden state $d_{st}^t$.

The dialogue context vector $d_o^t$ is then used by the user intent classifier
(Section \ref{sec:intent-classifier}) and user dialogue act classifier (Section
\ref{sec:user-act}). The slot tagger
(section \ref{sec:slot-tagger}) uses the dialogue context from previous turn
$d_o^{t-1}$, the system act vector $a^t$ and contextual token embeddings $u_o^t$
to generate refined contextual token embeddings $s_o^t$. These refined token
embeddings are then used to predict the slot tag for each token in the user
utterance.

The system dialogue acts and predicted slot tags are then used to update
the set of candidate values for each slot (Section \ref{sec:candidate-set}).
The candidate scorer (Section \ref{sec:candidate-scorer}) then uses the
previous dialogue state $D^{t-1}$, the dialogue context vector $d_o^t$ and
other features extracted from the current turn (indicated by blue arrows in
Figure \ref{fig:fig2}) to update the scores for all candidates in the candidate
set and outputs the updated dialogue state $D^t$. The following sections describe
these components in detail.

 \densesubsection{System Act Encoder}
\label{sec:system-act}
Previous turn system dialogue acts play an
important role in accurate semantic parsing of a user utterance. Each system
dialogue act contains an act type and optional slot and value parameters. The
dialogue acts are first encoded into binary vectors denoting the
presence of an act type. All dialogue acts which don't have any associated
parameters (e.g. \textit{greeting} and \textit{negate}) are encoded as a binary
indicator vector $a_{utt}^t$. Dialogue acts with just a slot $s$ as parameter
(e.g. \textit{request(date)}) are encoded as $a^t_{slot}(s)$, whereas acts
having a candidate value $c$ for a slot $s$ as parameter (e.g.
\textit{offer(time=7pm)}) are encoded as $a^t_{cand}(s, c)$. These binary
vectors are then combined using equations \ref{eqn:act-1}-\ref{eqn:act-4} to
obtain the combined system act representation $a^t$, which is used by other
units of dialogue encoder (as shown in Figure \ref{fig:fig2}). In these
equations, $e_s$ is a trainable slot embedding defined for each slot $s$.

\vspace{-10pt}
\begin{gather}
\label{eqn:act-1}
  {a}_{sc}^t(s) = a_{slot}^t(s) \oplus e_s \oplus \Sigma_c a_{cand}^t(s, c) \\
\label{eqn:act-2}
  {a'}_{sc}^t(s) = ReLU(W_{sc}^a \cdot {a}_{sc}^t(s) + b_{sc}^a) \\
\label{eqn:act-3}
  a_{usc}^t = \Big(\frac{1}{|S^t|}\sum_{s\in S^t} {a'}_{sc}^t(s)\Big) \oplus a_{utt}^t \\
\label{eqn:act-4}
  a^t = ReLU(W^a_{usc} \cdot a_{usc}^t + b^a_{usc})
\end{gather}

 \densesubsection{Utterance Encoder}
\label{sec:utterance-encoder}
The user utterance takes the tokens corresponding to the user utterance as
input. Special tokens \texttt{SOS} and \texttt{EOS} are added at the beginning
and end of the token list. Let
$x^t = \{x_m^t \in \mathbb{R}^{u_d}, \forall \, 0 \leq m < M^t\}$
denote the embedded representations of these tokens, where $M^t$ is the
number of tokens in the user utterance for turn $t$ (including \texttt{SOS} and
\texttt{EOS}).

We use a single layer bi-directional GRU recurrent neural network
\cite{grupaper} with state size $d_u$ and initial state set to 0, to encode the
user utterance. The first output of the user utterance encoder is
$u^t_e \in \mathbb{R}^{2d_u}$,
which is a compact representation of the entire user utterance, defined as
the concatenation of the final states of the two RNNs. The second output is
$u^t_o = \{u_{o,m}^t \in \mathbb{R}^{2d_u}, 0 \leq m < M^t$, which is the
embedded representation of each token conditioned on the entire utterance,
defined as the concatenation of outputs at each step of the forward and
backward RNNs.

 \densesubsection{State Encoder}
\label{sec:state-encoder}
The state encoder completes our hierarchical dialogue encoder. At turn $t$,
the state encoder generates $d_{o}^t$, which is an embedded representation of
the dialogue context until and including turn $t$. We implement the state
encoder using a unidirectional GRU RNN with each timestep corresponding to a
dialogue turn. As shown in Figure \ref{fig:fig2}, the dialogue encoder takes
$a^t \oplus u_e^t$ and its previous hidden state  $d_{st}^{t-1}$ as input and outputs
the updated hidden state $d_{st}^t$ and the encoded representation of the dialogue
context $d_o^t$ (which are the same in case of GRU).


 \densesubsection{User Intent Classification}
\label{sec:intent-classifier}
The user intent is used to identify the backend with which the dialogue
system should interact. We predict
the intents at each turn to allow user to switch intents during the dialogue.
However, we assume that a given user utterance can contain at-most one intent
and model intent prediction as a multi-class classification problem. At each
turn, the distribution over all intents is calculated as
\vspace{-5pt}
\begin{equation}
p_{i}^t = softmax(W_i \cdot d_{o}^t + b_i)
\end{equation}
where $\dim (p_{i}^t) = |I|$, $W_i \in \mathbb{R}^{d \times |I|}$ and
$b_i \in \mathbb{R}^{|I|}$, $I$ denoting the user intent vocabulary and
$d = \dim (d_o^t)$. During inference, we predict $argmax(p_{i}^t)$ as the intent
label for the utterance.

 \densesubsection{User Dialogue Act Classification}
\label{sec:user-act}
Dialogue acts are structured semantic representations of user utterances.
User dialogue acts are used by the dialogue manager in deciding
the next system action. We model user dialogue act classification as a multilabel classification
problem, to allow for the presence of more than one dialogue act in a turn
\cite{tur2011spoken}. At each turn, the probability for act $a$ is predicted as
\vspace{-5pt}
\begin{equation}
p_{a}^t = sigmoid(W_a \cdot d_{o}^t + b_a)
\end{equation}
where $\dim (p_a^t) = |A_u|$, $W_a \in \mathbb{R}^{d \times |A_u|}$,
$b_a \in \mathbb{R}^{|A_u|}$, $A_u$ is the user dialogue act vocabulary and
$d = \dim (d_o^t)$. For each act $\alpha$, $p_a^t(\alpha)$ is interpreted as
the probability of presence of $\alpha$ in turn $t$. During inference, all
dialogue acts with a probability greater than $t_u$ are predicted, where
$0 < t_u < 1.0$ is a hyperparameter tuned using the dev set.

 \densesubsection{Slot Tagging}
\label{sec:slot-tagger}
Slot tagging is the task of identifying the presence of values of different
slots in the user utterance. We use the IOB tagging
scheme (\citealt{tjong2000introduction}, see Figure \ref{fig:clu-example}) to assign a label to each token.
These labels are then used to extract the values
for different slots from the utterance.

The slot tagging network consists of a single-layer bidirectional LSTM RNN
\cite{lstmpaper}, which takes the contextual token embeddings $u_o^t$ generated
by the utterance encoder as input. It outputs refined token embeddings
$s_o^t = \{ s_{o, m}^t, \forall \, 0 \leq m < M^t \}$ for each token,
$M^t$ being the number of tokens in user utterance at turn $t$.

Models making use of dialogue context for LU
have been shown to achieve superior performance
\cite{chen2016end}. In our setup, the dialogue context vector
$d_o^{t-1}$ encodes all the preceding turns and the system act vector
$a^t$ encodes the system dialogue acts preceding the user
utterance. As shown in Figure \ref{fig:fig2}, $d_o^{t-1}$ is used to
initialize
\footnote{After projection to the appropriate dimension.}
the hidden state (cell states are initialized to zero) for the
forward and backward LSTM recurrent units in the slot tagger, while $a^t$ is fed as
input to the tagger by concatenating with each element
of $u_o^t$ as shown below. We use an LSTM instead of a GRU for
this layer since that resulted in better performance on the
validation set.

\vspace{-10pt}
\begin{gather}
    s^t_{in} = \{ u_{o, m}^t \oplus a^t, \forall \, 0 \leq m < M^t \} \\
    s^t_{e,bw}, s^t_{o,bw} = LSTM_{bw}(s^t_{in}) \\
    s^t_{e,fw}, s^t_{o,fw} = LSTM_{fw}(s^t_{in}) \\
    s^t_o = s^t_{o,bw} \oplus s^t_{o,fw}
\end{gather}

Let $S$ be the set of all slots in the dataset. We define a set of $2|S| + 1$
labels (one B- and I- label for each slot and a single O label)
for IOB tagging. The refined token embedding $s^t_{o, m}$ is used to predict
the distribution across all IOB labels for token at index $m$ as

\vspace{-10pt}
\begin{equation}
p_{s, m}^{t} = softmax(W_s \cdot s^t_{o, m} + b_s)
\end{equation}
where $\dim (p_{s, m}^{t}) = 2|S| + 1$,
$W_s \in \mathbb{R}^{d_s \times 2|S| + 1}$ and $b_s \in \mathbb{R}^{2|S| + 1}$,
$d_s = \dim (s^t_{o, m})$ is the output size of slot tagger LSTM. During
inference, we predict $argmax(p_{s, m}^{t})$ as the slot label for the $m^{th}$ token in the
user utterance in turn $t$.

 \densesubsection{Updating Candidate Set}
\label{sec:candidate-set}
A candidate set $C^t_s$ is defined as a set of values of a slot $s$ which have
been mentioned by either the user or the system till turn $t$.
\citealt{rastogi2017} proposed the use of candidate sets in DST for efficiently
handling slots with a large set of values. In their setup, the candidate set is
updated at every turn to include new values and discard old values when it
reaches its maximum capacity. The dialogue state is represented as a set of
distributions over value set $V_s^t = C_s^t \cup \{ \delta, \phi \}$
for each slot $s \in S^t$, where $\delta$ and $\phi$ are special values
\texttt{dontcare} (user is ok with any value for the slot) and \texttt{null}
(slot not specified yet) respectively, and $S^t$ is the set of all slots
that have been mentioned either by the user or the system till turn $t$.

Our model uses the same definition and update rules for candidate sets. At each
turn we use the predictions of the slot tagger (Section \ref{sec:slot-tagger})
and system acts which having slot and value parameters
to update the corresponding candidate sets. All candidate sets are padded with
dummy values for batching computations for all slots together. We keep track
of valid candidates by defining indicator features $m^{t}_v(s, c)$ for each
candidate, which take the value $1.0$ if candidate is valid or $0.0$ if not.

\begin{figure*}[t]
  \centering
  \includegraphics[width=\linewidth]{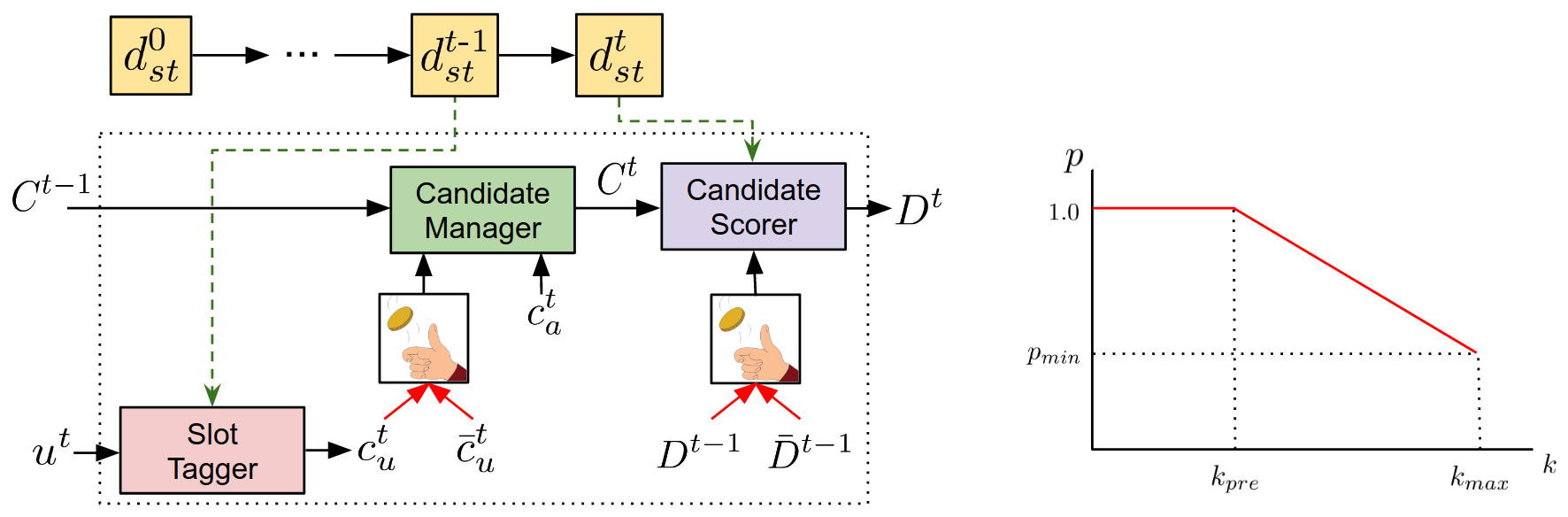}
  \caption{Illustration of scheduled sampling for training the candidate scorer.
        The left figure shows the two locations in our setup where
        we can perform scheduled sampling, while the plot on the right shows
        the variation of sampling probabilities $p_c$ and $p_D$
        with training step. See Section \ref{sec:scheduled} for details.}
  \label{fig:scheduled}
\end{figure*}

 \densesubsection{Candidate Scorer}
\label{sec:candidate-scorer}
The candidate scorer predicts the dialogue state by updating the distribution
over the value set $V_s^t$ for each slot $s \in S^t$. For this, we define three
intermediate features $r^t_{utt}$, $r^t_{slot}(s)$ and $r^t_{cand}(s, c)$.
$r_{utt}^t$ is shared across all value sets and is defined by equation
\ref{eqn:scorer-1}. $r^t_{slot}(s)$ is used to update scores for $V_s^t$ and is
defined by equation \ref{eqn:scorer-2}. Furthermore, $r^t_{cand}(s, c)$
is defined for each candidate $c \in C_s^t \subset V_s^t$ using equation
\ref{eqn:scorer-3} and contains all features that are associated to candidate
$c$ of slot $s$.

\vspace{-10pt}
\begin{gather}
\label{eqn:scorer-1}
  r_{utt}^t = d_o^t \oplus a_{utt}^t \\
\label{eqn:scorer-2}
  {r}_{slot}^t(s) = a_{slot}^t(s) \oplus [p^{t-1}_{\delta}(s), p^{t-1}_{\phi}(s)]
\end{gather}
\vspace{-0.3in}
\begin{equation}
\label{eqn:scorer-3}
\begin{aligned}
  {r}_{cand}^t(s, c) = & \, a_{cand}^t(s, c) \, \oplus \, [p^{t-1}_c(s)] \, \oplus \\
                       & [m^{t}_v(s, c), \, m^{t}_u(c)]
\end{aligned}
\end{equation}

In the above equations, $d_o^t$ is the dialogue context at turn $t$ output by
the state encoder (Section \ref{sec:state-encoder}), $a_{utt}^t$,
$a_{slot}^t(s)$ and $a_{cand}^t(s, c)$ are system act encodings generated by the
system act encoder (Section \ref{sec:system-act}), $p^{t-1}_{\delta}(s)$ and
$p^{t-1}_{\phi}(s)$ are the scores associated with \texttt{dontcare} and
\texttt{null} values for slot $s$ respectively. $p^{t-1}_c(s)$ is the score
associated with candidate $c$ of slot $s$ in the previous turn and is taken to
be $0$ if $c \not\in C_s^t$. $m^{t}_v(s, c)$ are variables indicating whether
a candidate is valid or padded (Section \ref{sec:candidate-scorer}). We define
another indicator feature $m^{t}_u(c)$ which takes the value $1.0$ if the
candidate is a substring of the user utterance in turn $t$ or $0.0$ otherwise.
This informs the candidate scorer which candidates have been mentioned most
recently by the user.

\vspace{-10pt}
\begin{gather}
\label{eqn:scorer-4}
  {r'}_{slot}^t(s) = r_{utt}^t \oplus {r}_{slot}^t(s) \\
\label{eqn:scorer-5}
  l^t_s(\delta) = {FF}^1_{cs}({r'}_{slot}^t(s)) \\
\label{eqn:scorer-6}
  l^t_s(c) = {FF}^2_{cs}({r'}_{slot}^t(s) \oplus {r}_{cand}^t(s, c)) \\
\label{eqn:scorer-7}
  p^t_s = softmax(l_s^t)
\end{gather}

Features used in Equations \ref{eqn:scorer-1}-\ref{eqn:scorer-3} are then used
to obtain the distribution over $V_s^t$ using Equations
\ref{eqn:scorer-4}-\ref{eqn:scorer-6}. In the above equations, $l^t_s(\delta)$ denotes the logit
for \texttt{dontcare} value for slot $s$, $l^t_s(c)$ denotes the logit for a
candidate $c \in C_s^t$ and $l^t_s(\phi)$ is a trainable parameter.
These logits are obtained by processing the
corresponding features using feed-forward neural networks $FF^1_{cs}$ and
$FF^2_{cs}$, each having one hidden layer. The output dimension of these
networks is 1 and the dimension of the hidden layer is taken to be half of the
input dimension. The logits are then normalized using softmax to get the
distribution $p^t_s$ over $V_s^t$.

 \densesection{Scheduled Sampling}
\label{sec:scheduled}

DST is a recurrent model which uses predictions from the previous turn. For
stability during training, ground truth predictions from the previous turn are
used. This causes a mismatch between training and inference behavior. We use
scheduled sampling ~\cite{bengio2015scheduled} to bridge this mismatch.
Scheduled sampling has been shown to achieve improved slot tagging performance
on single turn datasets ~\cite{liu2016joint}. Figure \ref{fig:scheduled} shows
our setup for scheduled sampling for DST, which is
carried out at two different locations - slot tags and dialogue
state.


The performance of slot
tagger is critical to DST because any slot value missed by
the slot tagger will not be added to the candidate set (unless it is tagged in
another utterance or present in any system act). To account for this, during training,
we sample between the ground truth slot tags ($\bar{c}_u^t$) and the predicted slot tags
($c_u^t$), training initially with $\bar{c}_u^t$ (i.e. with keep
probability $p_c = 1$)  but gradually reducing $p_c$ i.e. increasingly
replacing $\bar{c}_u^t$ with $c_u^t$. Using predicted slot tags
during training allows DST to train in
presence of noisy candidate sets.

During inference, the candidate scorer only has access to its own
predicted scores in the previous turn (Equations \ref{eqn:scorer-2} and
\ref{eqn:scorer-3}). To better mimic this setup during training, we start with
using ground truth previous scores taken from $\bar{D}^{t-1}$ (i.e. with keep
probability $p_D = 1$) and
gradually switch to $D^{t-1}$, the predicted previous scores, reducing $p_D$.

Both $p_c$ and $p_D$ vary as a function of the training step $k$, as shown
in the right part of Figure \ref{fig:scheduled}; only
ground truth slot tags and dialogue state are used for training i.e. $p_c$ and $p_D$ stay at 1.0
for the first $k_{pre}$ training steps,
and then decrease linearly as the ground truth slot tags and state are increasingly
replaced by model predictions during training.

 \densesection{Experiments}
\label{sec:experiments}
The major contributions of our work are two-fold. First, we hypothesize that
joint modeling of LU and DST results in a computationally efficient model with
fewer parameters without compromising performance. Second, we propose the use of
scheduled sampling to improve the robustness of DST during inference. To this
end, we conduct experiments across the following two setups.

\noindent
\\\textbf{Separate vs Joint LU-DST} - Figure \ref{fig:fig2} shows the joint LU-DST
setup where parameters in the utterance encoder and state encoder are shared
across LU tasks (intent classification, dialogue act classification and slot
tagging) and DST (candidate scoring). As baselines, we also conduct
experiments where LU and DST tasks use separate parameters for utterance and
state encoders.

\noindent
\\\textbf{Scheduled Sampling} - We conduct scheduled sampling (as described in
Section \ref{sec:scheduled}) experiments in four different setups.
\begin{enumerate}[noitemsep,topsep=0pt,leftmargin=12pt]
  \item \textit{None} - Ground truth slot tags ($\bar{c}_u^t$) and previous
    dialogue state ($\bar{D}^{t-1}$) are used
    for training.
  \item \textit{Tags} - Model samples between ground truth ($\bar{c}_u^t$) and predicted ($c_u^t$) slot tags, sticking to ground truth previous state.
  \item \textit{State} - Model samples between ground truth ($\bar{D}^{t-1}$) and predicted ($D^{t-1}$)
    previous state, sticking to ground truth slot tags.
  \item \textit{Both} - Model samples between $\bar{D}^{t-1}$ and $D^{t-1}$ as well as between $\bar{c}_u^t$ and $c_u^t$.
\end{enumerate}

In the last three setups, we start sampling from predictions only after
$k_{pre} = 0.3 \, k_{max}$ training steps, as shown in Figure \ref{fig:scheduled}.

\densesubsection{Evaluation Metrics}
We report user intent classification accuracy, F1 score for user
dialogue act classification, frame accuracy for slot tagging and
joint goal accuracy and slot F1 score for DST. During DST evaluation, we always use
the predicted slot values and the dialogue state in the previous turn.
Slot frame accuracy is defined as the fraction of turns for which all slot
labels are predicted correctly. Similarly, joint goal accuracy is the fraction
of turns for which the predicted and ground truth dialogue state match
for all slots. Since it is a stricter metric than DST slot F1, we use it as
the primary metric to identify the best set of parameters on the validation set.

\densesubsection{Datasets}
We evaluate our approaches on two datasets:
\begin{itemize}[noitemsep, topsep=2pt, leftmargin=12pt]
\item \textbf{Simulated Dialogues}\footnote{Dataset available at
\href{http://github.com/google-research-datasets/simulated-dialogue/}{http://github.com/google-research-datasets/simulated-dialogue/}}
- The dataset, described in ~\citealt{M2M-overnight}, contains dialogues from restaurant (Sim-R) and
movie (Sim-M) domains across three intents. A challenging aspect of this dataset is
the prevalence of OOV entities e.g. only 13\% of the
movie names in the dev/test sets also occur in the training data.
\item \textbf{DSTC2} - We use the top ASR hypothesis and system dialogue acts as
inputs. Dialogue act labels are obtained from top SLU hypothesis and state labels
for requestable slots. DS labels are obtained from state labels for
informable slots. We use a semantic dictionary ~\cite{henderson2014} to obtain
ground truth slot tags. We also use the semantic dictionary to
canonicalize the candidate values since the slot values in the dialogue state
come from a fixed set in the DSTC2 dialogues and may be different from those
present in the user utterance.
\end{itemize}

\densesubsection{Training}
\label{sec:training}
We use sigmoid cross entropy loss for dialogue act classification and softmax
cross entropy loss for all other tasks. During training, we minimize the sum of
all task losses using ADAM optimizer \cite{kingma2014adam}, for 100k training
steps with batches of 10 dialogues each. We used grid-search to identify the
best hyperparameter values (sampled within specified range) for learning rate
(0.0001 - 0.005) and token embedding dimension (50 - 200). For scheduled
sampling experiments, the minimum keep rate i.e. $p_{min}$ is varied between
0.1 - 0.9 with
linear decay. The layer sizes for the utterance encoder and slot tagger are set
equal to the token embedding dimension, and that of the state encoder to half
this dimension.

\textbf{Slot Value dropout} - To make the model robust to OOV tokens arising
from new entities not present in the training set, we randomly replace
slot value tokens in the user utterance with a special OOV token with a
probability that linearly increases from 0.0 to 0.4 during training.

\densesection{Results and Discussion}
Table \ref{tab:experiments} shows our results
across the two setups described in
Section \ref{sec:experiments}, for the Simulated Dialogues datasets. For
Sim-R + Sim-M, we observe
that the joint LU-DST model with scheduled sampling (SS) on both slot tags and dialogue state
performs the best, with a joint goal accuracy of 73.8\% overall,
while the best separate model gets a joint goal accuracy of 71.9\%, using SS
only for slot tags. Even for the no-SS baselines, the joint model performs
comparably to the separate model (joint goal accuracies of 68.6\%
and 68.7\% respectively), indicating that sharing
results in a more efficient model with fewer parameters, without compromising
overall performance. For each SS configuration, our results
comparing separate and joint modeling are statistically significant, as
determined by the McNemar's test with $p < 0.05$.

\begin{table*}[t]
  \caption{Experiments and results on test set with variants of scheduled sampling
           on separate and joint LU-DST models, when trained on Sim-M + Sim-R.}
  \label{tab:experiments}
  \centering
  \hspace{-19pt}
  \begin{tabularx}{1\textwidth}{l l | c c | c c | c c | c c | c c}
    \tabhline
    \textbf{Eval Set} & \textbf{SS} &  \multicolumn{2}{c}{\textbf{Intent}} & \multicolumn{2}{c}{\textbf{Dialogue Act}} & \multicolumn{2}{c}{\textbf{Slot Frame}} & \multicolumn{2}{c}{\textbf{Joint Goal}} & \multicolumn{2}{c}{\textbf{DST Slot}} \\
    & \textbf{Setup}  & \multicolumn{2}{c}{\textbf{Accuracy}} & \multicolumn{2}{c}{\textbf{F1 Score}} & \multicolumn{2}{c}{\textbf{Accuracy}} & \multicolumn{2}{c}{\textbf{Accuracy}} & \multicolumn{2}{c}{\textbf{F1 Score}} \tabhline
    & & Sep & Joint & Sep & Joint & Sep & Joint & Sep & Joint & Sep & Joint \tabhline
    Sim-R & None        & 0.999 & 0.997 & 0.956 & 0.935 & 0.924 & 0.919 & 0.850 & 0.846 & 0.951 & 0.952\\
          & Tags       & 0.998 & 0.998 & 0.936 & 0.957 & 0.917 & 0.922 & 0.805 & 0.871 & 0.936 & 0.962\\
          & State       & 0.999 & 0.998 & 0.931 & 0.939 & 0.919 & 0.920 & 0.829 & 0.852 & 0.935 & 0.951\\
          & Both        & 0.994 & 0.998 & 0.948 & 0.919 & 0.917 & 0.916 & 0.829 & 0.849 & 0.942 & 0.953 \tabhline

    Sim-M & None        & 0.991 & 0.993 & 0.966 & 0.966 & 0.801 & 0.800 & 0.276 & 0.283 & 0.806 & 0.817\\
          & Tags       & 0.993 & 0.994 & 0.970 & 0.967 & 0.895 & 0.801 & 0.504 & 0.262 & 0.839 & 0.805\\
          & State       & 0.996 & 0.970 & 0.964 & 0.955 & 0.848 & 0.799 & 0.384 & 0.266 & 0.803 & 0.797\\
          & Both        & 0.989 & 0.996 & 0.970 & 0.959 & 0.887 & 0.860 & 0.438 & 0.460 & 0.805 & 0.845 \tabhline

    Sim-R + & None        & 0.996 & 0.996 & 0.959 & 0.944 & 0.890 & 0.885 & 0.687 & 0.686 & 0.902 & 0.906\\
    Sim-M   & Tags       & 0.996 & 0.997 & 0.946 & 0.960 & 0.910 & 0.888 & 0.719 & 0.698 & 0.902 & 0.905\\
            & State       & 0.996 & 0.990 & 0.940 & 0.943 & 0.899 & 0.886 & 0.702 & 0.683 & 0.897 & 0.899\\
            & Both        & 0.993 & 0.997 & 0.954 & 0.931 & 0.909 & 0.900 & 0.717 & \textbf{0.738} & 0.894 & \textbf{0.915}  \tabhline

  \end{tabularx}
\end{table*}


On the Sim-R dataset, the best
joint model obtains a joint goal accuracy of 87.1\%, while the best separate model
obtains 85.0\%. However, we observe a significant drop in joint goal accuracy for the Sim-M dataset
for both the joint model and the separate model as compared to Sim-R. This can partly be attributed
to the Sim-M dataset being much smaller than Sim-R (384 training dialogues as opposed to
1116) and that the high OOV rate of the \textit{movie} slot in Sim-M makes
slot tagging performance more crucial for Sim-M. While SS does
gently bridge the gap between training and testing conditions, its gains are
obscured in this scenario possibly since it is very hard for DST to recover
from a slot value being completely missed by LU, even when aided by SS.

For the two datasets, we also observe a significant difference between the slot
frame accuracy and joint goal accuracy.  This is because an LU error penalizes
the slot frame accuracy for a single turn, whereas an error in dialogue state
propagates through all the successive turns, thereby drastically reducing the joint goal
accuracy. This gap is even more pronounced for Sim-M because of the poor
performace of slot tagger on \textit{movie} slot, which is often mentioned by
the user in the beginning of the dialogue. The relatively high values of overall
DST slot F1 for Sim-M for all experiments also corroborates this observation.

\vspace{-7pt}
\begin{table}[h]
  \caption{Reported joint goal accuracy of model variants on the DSTC2 test set.}
  \label{tab:dstc2-experiments}
  \centering
  \vspace{7pt}
  \begin{tabular}{| l | c c |}
    \hline
    \textbf{Model} & \textbf{Separate} & \textbf{Joint} \tabhline
    No SS & 0.661 & 0.650 \\
    Tags only SS  & 0.655 & \textbf{0.670} \\
    State only SS & 0.661 & 0.660 \\
    Tags + State SS & 0.656 & 0.658 \\\hline
    \citealt{BingLiu2017} & - & 0.73 \\
    \citealt{mrkvsic2017neural} & - & 0.734 \\\hline
  \end{tabular}
\end{table}

Table \ref{tab:dstc2-experiments} shows our results on the DSTC2 dataset, which
contains dialogues in the restaurant domain. The joint model gets a joint goal
accuracy of 65.0\% on the test set, which goes up to 67.0\% with SS on slot tags.
Approaches like NBT ~\cite{mrkvsic2017neural} or Hierarchical RNN
~\cite{BingLiu2017} are better suited for such datasets, where the set
of all slot values is already known, thus eliminating the need for slot tagging.
On the other hand, our setup uses slot tagging for candidate generation, which
allows it to scale to OOV entities and scalably handle slots with a large or
unbounded set of possible values, at the cost of performance.

Analyzing results for scheduled sampling, we observe that for almost
all combinations of metrics, datasets and joint/separate model configurations, the best
result is obtained using a model trained with some SS variant. For instance, for Sim-M,
SS over slot tags and state increases joint goal accuracy significantly from 28.3\%
to 46.0\% for joint model. SS on slot tags
helps the most with Sim-R and DSTC2: our two datasets with the
most data, and low OOV rates, while SS on both slot tags and dialogue state helps
more on the smaller Sim-M. In addition, we also found that slot value dropout (Section \ref{sec:training}), improves LU as well as DST results
consistently. We omit the results without this technique for brevity.

\densesection{Conclusions}
In this work, we present a joint model for language understanding (LU) and
dialogue state tracking (DST), which is computationally efficient by way of
sharing feature extraction layers between LU and DST, while achieving
an accuracy comparable to modeling them separately across multiple tasks.
We also demonstrate the effectiveness of scheduled sampling on LU outputs
and previous dialogue state as an effective way to simulate inference-time
conditions during training for DST, and make the model more
robust to errors.

\bibliography{acl2018}
\bibliographystyle{acl_natbib}

\end{document}